%% file: main.tex
\documentclass[10pt, a4paper]{article}
\usepackage{lrec-coling2024} 

\renewcommand{\rmdefault}{phv} 

\usepackage{booktabs}
\usepackage{tablefootnote}

\usepackage{natbib}
\usepackage{multibib}
\makeatletter
\def\@mb@citenamelist{cite,citep,citet,citealp,citealt,citepalias,citetalias}
\makeatother

\usepackage{graphicx}
\usepackage{tabularx}
\usepackage{soul}
\usepackage{listings}
\usepackage{todonotes}
\usepackage{multirow}
\usepackage{enumitem}
\usepackage{colortbl}
\usepackage{xcolor}
\usepackage{xstring}
\usepackage{lipsum}
\usepackage{threeparttable}
\usepackage{makecell}

\definecolor{Maroon}{rgb}{0.5, 0.0, 0.0}
\colorlet{color0}{Maroon!10} 
\colorlet{color1}{Maroon!20}
\colorlet{color2}{Maroon!20}
\colorlet{color3}{Maroon!30}
\colorlet{color4}{Maroon!30}
\colorlet{color5}{Maroon!40}
\colorlet{color6}{Maroon!50}
\colorlet{color7}{Maroon!60}
\colorlet{color8}{Maroon!60}
\colorlet{color9}{Maroon!70}
\colorlet{color10}{Maroon!70}
\colorlet{color11}{Maroon!70}
\colorlet{color12}{Maroon!75}
\colorlet{color13}{Maroon!80}
\colorlet{color14}{Maroon!85}
\colorlet{color15}{Maroon!90}

\usepackage{titlesec}
\titleformat{\section}{\normalfont\large\bfseries\center}{\thesection.}{1em}{}
\titleformat{\subsection}{\normalfont\SmallTitleFont\bfseries\raggedright}{\thesubsection.}{1em}{}
\titleformat{\subsubsection}{\normalfont\normalsize\bfseries\raggedright}{\thesubsubsection.}{1em}{}
\renewcommand\thesection{\arabic{section}}
\renewcommand\thesubsection{\thesection.\arabic{subsection}}
\renewcommand\thesubsubsection{\thesubsection.\arabic{subsubsection}}

\usepackage{hyperref}
\usepackage{cleveref}
\usepackage{arydshln}
\usepackage[nolist,nohyperlinks]{acronym}

\definecolor{darkblue}{rgb}{0, 0, 0.5}
\hypersetup{colorlinks=true, citecolor=darkblue, linkcolor=darkblue, urlcolor=darkblue}

\newcommand{\datasetname}{PECC}
\acrodef{AoC}[AoC]{\emph{Advent of Code}}

\title{\datasetname{}: Problem Extraction and Coding Challenges}

\name{Patrick Haller, Jonas Golde, Alan Akbik}

\address{Humboldt-Universität zu Berlin \\
        \{patrick.haller.1, jonas.golde, alan.akbik\}@hu-berlin.de\\}

\abstract{Recent advancements in large language models (LLMs) have showcased their exceptional abilities across various tasks, such as code generation, problem-solving and reasoning. Existing benchmarks evaluate tasks in isolation, yet the extent to which LLMs can understand prose-style tasks, identify the underlying problems, and then generate appropriate code solutions is still unexplored. Addressing this gap, we introduce \datasetname{}, a novel benchmark derived from Advent Of Code (AoC) challenges and Project Euler, including 2396 problems. Unlike conventional benchmarks, \datasetname{} requires LLMs to interpret narrative-embedded problems, extract requirements, and generate executable code. A key feature of our dataset is the complexity added by natural language prompting in chat-based evaluations, mirroring real-world instruction ambiguities. Results show varying model performance between narrative and neutral problems, with specific challenges in the Euler math-based subset with GPT-3.5-Turbo passing 50\% of the AoC challenges and only 8\% on the Euler problems. By probing the limits of LLMs' capabilities, our benchmark provides a framework to monitor and assess the subsequent progress of LLMs as a universal problem solver.
\\ \newline  }

\begin{document}

\maketitleabstract

\section{Introduction}

Large language models (LLMs) have demonstrated remarkable abilities in diverse generation tasks, spanning text and beyond. Consequently, they have become reliable tools for code generation, reducing barriers for entry-level engineers and supporting experienced programmers. Researchers have devised instruction-based benchmarks to assess the progress of code generation abilities in new language models \citep{2021_c24cd76e, hendrycks2021measuring, lai2022ds1000}. These benchmarks prompt a given LLM to generate executable code for specified problems. Despite the isolation of these benchmarks from other LLM advancements, such as enhanced reading comprehension or problem abstraction \citep{wang2019glue,reddy2019coqa,liu2020logiqa}, integrating all these skills could transform LLMs into universal problem-solving tools. Thus, how well LLMs can combine their abilities to (\textit{1}) understand prose style problems, (\textit{2}) identify solution requirements, and (\textit{3}) translate them into executable code remains mostly unexplored.

\begin{figure}[ht!]
    \centering
    \includegraphics[scale=0.36]{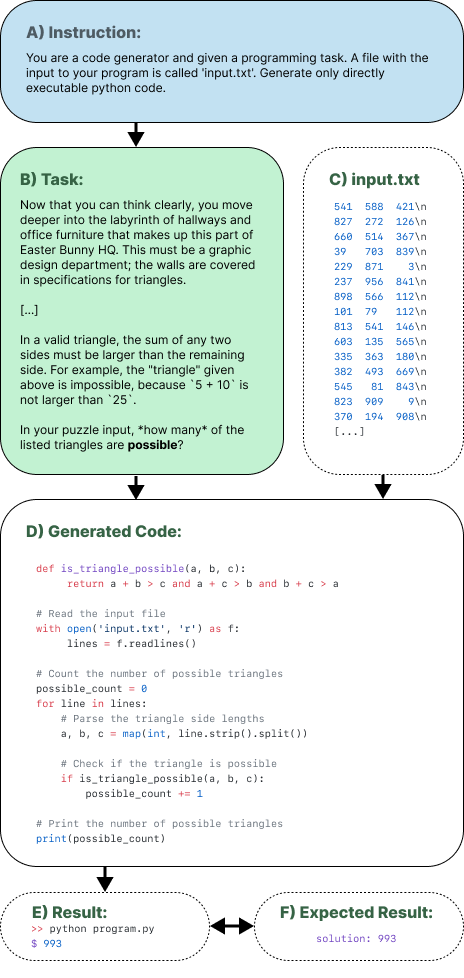}
    \caption{A schematic representation of the code generation and assertion process.}
    \label{fig:example}
\end{figure}

\input{tables/comparison}

The illustration in Figure~\ref{fig:example} delineates a systematic process of code generation and assertion that tests named abilities in an automated manner. The initial step (A) involves instructing the language model, setting the contextual foundation, and defining task requirements. Subsequently, the model receives a problem statement (B) for which it needs to generate a coding solution. The next step involves the model generating executable Python code (D) based on the provided task and input data (C). Execution of the generated code yields a result (E), which is then compared with an expected result (F). Directional arrows indicate the flow from instruction through to result-assertion. Recognizing this systematic process emphasizes the need for benchmarks that can truly test these intricate steps, especially when considering real-world applications where narrative understanding is pivotal.

In this paper, we introduce \datasetname{}, an extensive benchmark centered on code generation from narrative-embedded problem descriptions. Unlike prior benchmarks that evaluate code generation using specific instructions, our dataset requires models to comprehend, extract requirements, and produce the essential code for problem-solving. This approach necessitates syntactically accurate programs and demands reading comprehension skills to derive the desired solution.

Our dataset comprises 2,396 problems spanning different levels of difficulty. We use the problems from the annual Advent Of Code\footnote{https://adventofcode.com/} challenges and the online platform Project Euler\footnote{https://projecteuler.net/} as our source dataset. The challenges are presented in either prose style or neutral formats. We transform each dataset into the opposing writing style, enabling us to evaluate problem abstraction across different formulations, namely, neutral versus narrative formulation.

If a model were to perform well on \datasetname{}, it signifies the models' proficiency in accurately comprehending natural language specifications and generating correct code. This accomplishment relies on the model's ability to employ appropriate data structures and algorithms to solve the encapsulated problem. Additionally, the dataset allows us to assess the models' sequential problem-solving skills in real-world coding scenarios. This evaluation occurs when solutions to one problem are prerequisites for solving a subsequent problem, as with the Advent Of Code (AoC) split. In AoC, each day consists of two challenges, of which the second challenge can only be tackled using the solution derived from the first problem.

We summarize the contribution of this paper as follows: 
\begin{enumerate}
\item We introduce the construction process of \datasetname{}, a novel benchmark designed to evaluate LLMs in prose-style coding challenges.
\item We evaluate state-of-the-art language models and show that although they perform well on simple tasks, their performance drops significantly as the complexity of the task increases.
\item We thoroughly analyze performances, program errors, and different prompting schemes across all problem types and difficulty levels within \datasetname{}.
\item We will provide the dataset and evaluation framework to the research community at \url{https://github.com/HallerPatrick/pecc}. The pipeline supports multiple prominent LLM providers and local hosting for inference.
\end{enumerate}
\section{\datasetname{} Dataset}

The \datasetname{} dataset leverages two prominent resources: The Advent of Code (AoC) challenges and Project Euler.

\subsection{Dataset Construction}
\begin{description}[style=unboxed,leftmargin=0cm]
\item[Advent of Code.] AoC, an annual online event, unfolds daily coding challenges throughout December, each embedded within a festive narrative. It presents complex problems, enabling coders to refine their problem-solving skills within a playful, story-driven context.
\item[Project Euler.] On the other hand, Project Euler, an online platform, curates mathematical and computational problems that necessitate a blend of mathematical understanding and programming skills for resolution. It fosters learning and problem-solving within a community-driven environment, featuring progressively escalating difficulty levels. Difficulty levels are increasing in steps of 5 from 0 to 100.
\end{description}
\begin{figure*}
    \centering
    \includegraphics[scale=0.5]{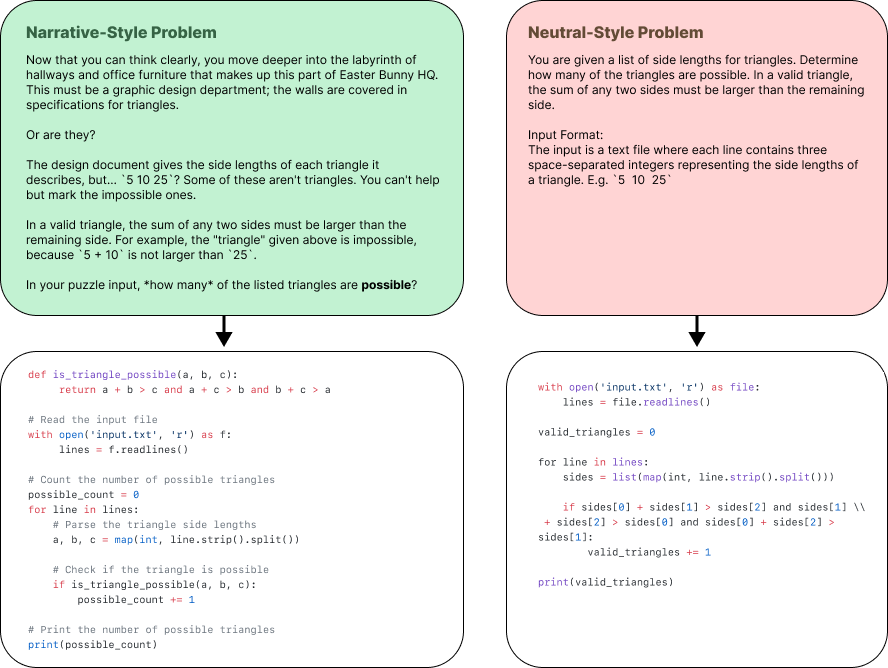}
    \caption{Contrasting Problem Descriptions from AoC. The left illustrates a narrative-style problem, rich in story and context, while the right presents a neutral-style, succinctly distilled version with their respective generated solutions with GPT-3.5-Turbo. We observe that the generated code is more concise for neutrally formulated problems, while the solutions for narrative problems tend to model the story more.}
    \label{fig:problem_and_solution}
\end{figure*}

Problem presentation and verbosity in AoC and Project Euler stand in striking contrast to each other. While AoC problems are narrative-based, often enveloped in a story with elongated problem descriptions, Project Euler offers precise, succinct ones. This distinction reflects the divergent approaches of the two platforms in engaging their audience and elucidating problems. These problem verbosities provide a rich foundation for formulating the \datasetname{} dataset.

\subsection{Augmenting AoC and Euler Problem Styles}
We augment complementary datasets to explore how different problem formulations - narrative-style or neutral-style - affect problem-solving capabilities. From the concise descriptions of Project Euler, we generate problems with a narrative twist, adding a story-driven context. Conversely, we transform detailed AoC problems into neutral-toned challenges to resemble the directness of Project Euler. These augmentations yield diverse problem presentations to test a language model's adaptability to varying problem contexts. The original datasets encompass 392 problems from AoC and 806 problems from Project Euler, totaling 1,198 problems. After creating complementary data points, the dataset comprises 2,352 problems, each with its corresponding solution, typically an integer, and, in the case of AoC, the actual input required to solve the challenges.
Figure \ref{fig:problem_and_solution} illustrates complementary problems and their generated solutions.
We denote the original sources as \emph{aoc} and \emph{euler} for the Advent of Code challenges and the Euler Project, respectively. The generated complements as \emph{aoc-concise} and \emph{euler-stories}.

A common assumption might be that for a language model to generate a complementary description of a given problem, it must be able to solve the problem itself. However, our quality checks showed that the capacity to describe or rephrase a problem does not necessarily imply a profound understanding or the ability to solve it.


\subsection{Single- and Multi-Turn Prompting}
Our dataset employs two distinct prompt formulations. We use a (\textit{1}) instruction-based format for the Euler subset to direct models to create executable Python programs and return answers via standard output. Conversely, we employ a (\textit{2}) chat-based format for the AoC subset due to its two-part problem structure. We instruct models to generate directly executable Python code, with the addition of loading input from an "input.txt" file, for which models also need to generate the relevant code, ensuring a comprehensive engagement with the problem-solving scenario. We repeat this procedure for the second part of the AoC challenge, including the previous conversation history. Every AoC challenge begins with a self-contained first part and can use an instruction-based format by including the "input.txt" file. In alignment with existing benchmarks, we employ a zero-shot format for \datasetname{}, but all problems come without few-shot examples.

\section{Experiments}

\subsection{Experimental Setup}

We evaluate various language models on \datasetname{} to investigate the extent to which these models can harness their inherent capabilities to solve complex coding tasks considering different problem formulations. We
select GPT-3.5-turbo-16k (gpt-3.5-turbo)~\citep{OpenAI2023GPT4TR} from OpenAI and VertexAI's PALM 2 (chat-bison)~\citep{anil2023palm}, Codey (codechat-bison) and Claude 3 Haiku (claude-haiku)~\citep{claude-3}, as these models are among the most capable in terms of natural language processing and code-generation tasks.
 Additionally, we tested several different open-source language models, that were instruction and chat-fine tuned:  
 \citep{luo2023wizardcoder,jiang2023mistral,jiang2024mixtral,llama3modelcard,abdin2024phi3}
The latter part of our study focused on comparing the performance of these models, which is essential for discerning their relative strengths and weaknesses in code generation contexts.

Our evaluation pipeline builds upon the Langchain~\citep{langchain} library, which facilitates chat and instruction-based formats for interacting with the models. The generated source code is executed in an isolated Python environment, ensuring a secure and controlled execution process. 

\input{tables/result_table}
\subsection{Qualitative Error Mode Analysis}
During code execution, we extract and categorize errors encountered into five distinct types:
\begin{enumerate}
    \item \textbf{Syntax Error}: Arises when LLM generates no or only partial Python source code, or the Python interpreter returns a SyntaxError. 
    \item \textbf{Runtime Error}: Occurs when a program, despite being syntactically correct, aborts due to an error during execution, such as IndexErrors, KeyError, and NameError. 
    \item \textbf{Timeout Error}: Triggered when a program's runtime exceeds a set threshold of 60 seconds.
    \item \textbf{Wrong Output}: Specified for when a program exits successfully but yields an output differing from the expected result.
    \item \textbf{Part1 Failed}: Occurs when Part 1 of the AoC challenges did not pass.
\end{enumerate}

This structured error categorization aids in a granular analysis of the models' performance, pinpointing the areas of strength and the potential avenues for improvement.


\subsection{Metrics}

\textbf{Pass@k.} The Pass@k metric evaluates model code synthesis by checking if at least one of the top 'k' generated code snippets passes predefined unit tests. Utilized in frameworks like CodeEval, it assesses how well models translate natural language prompts into accurate, executable code. It quantitatively measures a model's capability to generate functionally valid code. In our evaluation, functionally valid code is defined as code able to execute and return the expected result via standard out. Adopting of Pass@k allows for a nuanced understanding of model performance, especially in scenarios where multiple code solutions might be plausible.

\textbf{Pass@k + Difficulty.} Introducing an additional difficulty term to Pass@k, termed Pass@k-Difficulty, enables a discriminative evaluation of a model’s ability to tackle problems of varying difficulty, furnishing a detailed understanding of the model’s strengths and weaknesses. The difficulty weight is determined based on the percentage of participants successfully solving a given problem, providing a real-world reflection of the problem's complexity. Pass@k-Difficulty is realized by introducing a weighting factor based on problem difficulty to the existing Pass@k metric formula accounting for problem difficulty, with higher weights accorded to more complex problems, thus presenting a more accurate reflection of a model’s performance in challenging scenarios. We collected user statistics about the ground population and the number of participants who solved each problem from their respective website \footnote{Statistics for AoC accessible at \url{https://adventofcode.com/2022/stats}  and Euler Project at \url{https://projecteuler.net/problem_analysis}}

\subsection{Prompt Formulation}

Further, we compare models' performance by employing varied prompt formulations. While generating and executing Python code for problem-solving presents a formidable challenge, this experiment centers on determining whether LLMs can effectively tackle these challenges by leveraging their inherent world knowledge, defined as their intrinsic aptitude to solve logical or mathematical problems.

Furthermore, we introduce an additional layer of complexity by compelling the models to employ a systematic chain-of-thought procedure, requiring them to justify and elucidate their answers comprehensively.

Summarized, we have delineated this experiment into two distinct contexts: the first, where LLMs respond to challenges relying solely on their innate world knowledge, termed \emph{answering}; and the second, where LLMs are tasked with justifying their solutions through a \emph{chain-of-thought} procedure, denoted as \emph{answering + chain-of-thought}.

\section{Results}

\subsection{Output Evaluation}
We show the main results of our evaluation in Table \ref{tab:output_evaluation}. The presented table delineates a comparative evaluation of several language models across distinct subsets of the \datasetname{} dataset, observing the performance at $k=1$ and $k=3$ for the respective problem subsets. The models evaluated include proprietary models from OpenAI and VertexAI, along with an open-source instruction and chat-fine-tuned models.

\noindent
\textbf{Multi-sampling increases the likelihood of correct solutions.}
It is discernible that employing a multi-sampling approach with $k=3$ generally improves the Pass@k scores compared to when $k=1$, which is single sampling. This improvement is particularly noticeable in the Advent of Code (AoC) subset for the gpt-3.5-turbo and claude-haiku models. The increment in scores from $k=1$ to $k=3$ suggests that providing models with multiple attempts to solve a problem enhances the likelihood of generating a correct solution. However, multi-sampling does not always guarantee better results. If a model cannot inherently solve a given problem, as observed with the mistral-instruct model, then multiple passes will not necessarily lead to a better score. The efficacy of multi-sampling is contingent on the model's base proficiency in addressing the task at hand.

\noindent
\textbf{Narratives can aid or obstruct models.}
The differing performances on AoC and Euler datasets reveal that narratives can be a double-edged sword. In the evaluation, the story-driven subset AoC proved better suited for models than its neutrally formulated counterpart. These narratives in AoC provide vital context, enhancing the model's capability to decode and solve problems. Conversely, introducing narratives led to a drop in performance for the Project Euler (Euler) subset, which consists of precise mathematical problems. This reduction suggests that adding stories to Euler introduces ambiguity, complicating the problems and reducing model effectiveness. This stark difference in performance across the two datasets underscores a potential training bias in models: they may excel in narrative-rich environments like AoC but struggle with the precision-driven challenges inherent to Euler.

\noindent
\textbf{LLMs generally fall short on complicated challenges.}
The evaluations underscore the challenges inherent in complex coding problems, particularly evident in the Euler subset. gpt-3.5-turbo and claude-haiku emerge as relatively strong performers among the lot, yet the results also highlight a significant room for improvement across all models. The assessment subtly hints at the prospective merits of fine-tuning or employing models specialized in code synthesis for enhanced performance in such tasks.

\input{tables/error_types}
\textbf{Potential Influence of Pre-training Contamination.}
Given the popularity and prominence of the AoC problems, there is a conceivable risk of pre-training contamination for language models. It is plausible that many of these models, during their training phase, have been exposed to myriad solutions to the AoC problems publicly available in Python. This exposure raises the question of authenticity in the model's response: Is the model genuinely attempting to solve the problem on its merit, or is it merely regurgitating a previously encountered solution? Determining the extent of this influence is challenging. We cannot unequivocally discern whether a model produces a solution based on its problem-solving capability or relies on a recollection of previously seen solutions. This potential for contamination underscores the complexity of evaluating language models on tasks where the training data might overlap with evaluation benchmarks.

\subsection{Error Analysis}

Table \ref{tab:error_types} presents meticulous error analysis across all evaluated models and dataset subsets, categorizing errors into Runtime Error, Wrong Output, Timeout, and Syntax Error. It is evident across all models that Syntax Errors occur less frequently, indicating a reasonable adherence to Python's syntactic rules. However, models like gpt-3.5-turbo exhibit significant Runtime and Wrong Output errors, especially in mathematically intensive datasets like Euler and Euler Stories, suggesting challenges in logically or mathematically solving the problems despite syntactical correctness.
A consistent pattern of errors is observed across models like Chat-Bison, CodeChat-Bison, and Wizard, where Runtime Errors are predominant, indicating struggles with the logical or algorithmic aspects of the problems. Notably, the Wizard model shows a relatively higher occurrence of Syntax Errors in Euler and Euler Stories datasets, hinting at challenges in translating mathematical or narrative elements into correct code.

\subsection{Coding vs. World Knowledge}

\begin{figure*}[h]
    \centering
    \includegraphics[width=\textwidth]{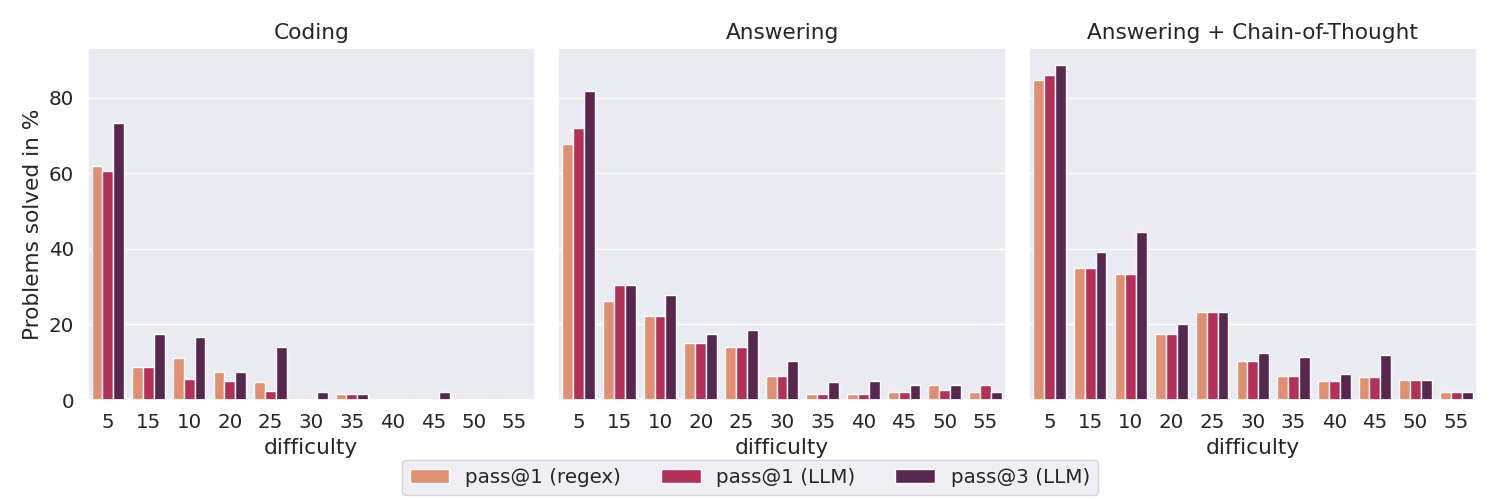}
    \caption{We present the percentage of Euler problems solved using gpt-3.5-turbo, categorized by their difficulty levels. For the easiest category, correct solutions were obtained in the range of \textasciitilde~60\% to \textasciitilde~80\% for coding and answering using chain-of-thought, respectively. However, as the difficulty level increases, the success rate drops rapidly. We do not report scores for difficulty levels higher than 55, as gpt-3.5-turbo did not provide any correct answers in those cases.}
    \label{fig:euler_scores}
\end{figure*}

In this ablation experiment, we compare the performance of solving Euler problems by code and compare it to answers directly obtained through prompts. The results are shown in \Cref{fig:euler_scores}.

\noindent
\textbf{Coding is more difficult than answering.} We initially noticed that the coding scores were lower than when the LLM generated the answers. In the most straightforward scenarios, coding performance dropped by 20pp. compared to answering settings. This finding suggests that GPT can effectively depend on its reasoning ability. Additionally, we observed that coding tasks were rarely solved when the difficulty level reached 30.

\noindent
\textbf{Chain-of-Thought improves performance.} We discovered that prompting GPT to provide reasons for its solutions enhances performance across all difficulty levels. In the most straightforward problem setting, gpt-3.5-turbo solves > 80\% problems, and this technique also enables it to solve some of the more challenging problems. Compared to answering without chain-of-thought, using chain-of-thought results in a notable improvement of +13.8 pp. on difficulty level 10, for instance.

\subsection{Problem Difficulty}

\begin{figure*}[h]
    \centering
    \includegraphics[width=\textwidth]{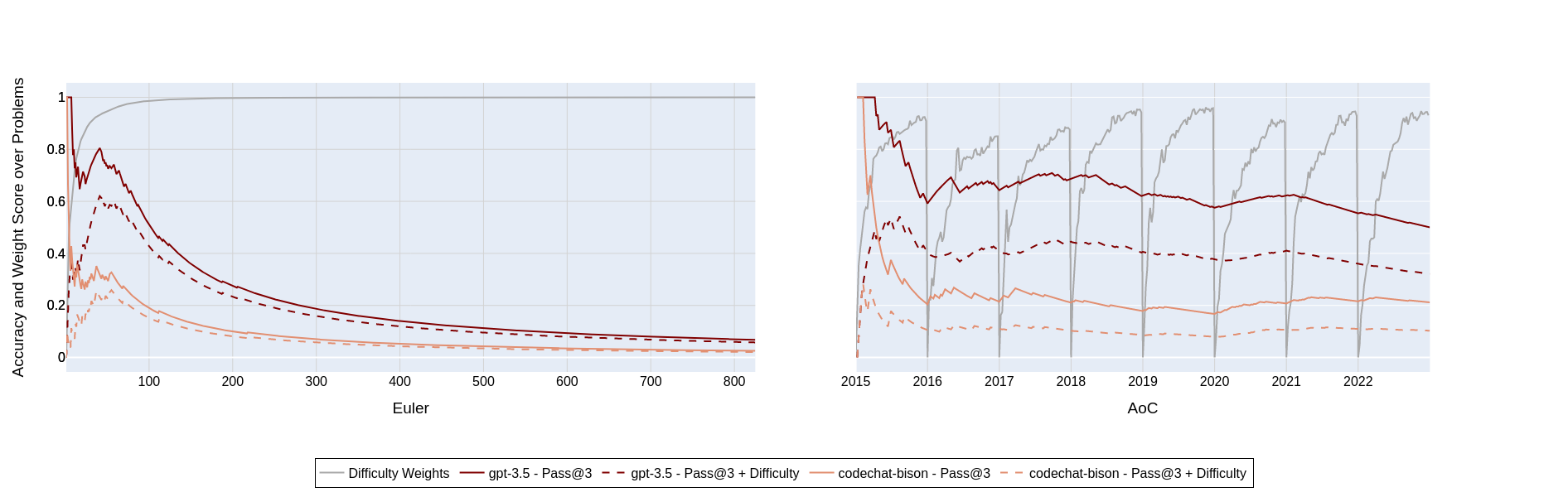}
    \caption{Comparing the accuracy with Pass@3 and Pass@3 + Difficulty for gpt-3.5-turbo and codechat-bison over the \textit{Euler} and \textit{AoC} subsets.}
    \label{fig:comp_weight}
\end{figure*}

Figure \ref{fig:comp_weight} shows the accuracy development overall problems for the AoC and Euler subsets. The Euler problems are sorted by difficulty, while the AoC problems are sorted by year and day. The gray line plot describes the difficulty weight over each problem. The plot contains the progressive accuracy over Pass@3 and Pass@3-Difficulty.
In the comparative evaluation of gpt-3.5-turbo and codechat-bison, clear distinctions were observed across the Euler and AoC subsets of the dataset. The Euler subset displayed a more predictable progression, with the problem difficulty increasing from left to right. This systematic rise in difficulty aligns with the decreasing accuracy trends of both models.

The AoC challenges show a reoccurring pattern of difficulty levels for each year. Historically, AoC problems tend to begin easier at the start of the year and become more challenging towards the year-end. This periodic nature of difficulty is captured in the multiple spikes observed in the gray plot. These spikes represent challenges that are notably harder than their preceding problems. Such a non-linear distribution can lead to varied performance from the models, where they might excel in some segments and face challenges in others, even within a short span of problems.

Considering these nuances, it becomes apparent that a one-size-fits-all evaluation metric, like raw accuracy, might not capture the intricacies of a model's performance, especially in datasets like AoC with non-uniform difficulty. By incorporating problem difficulty as a weighting factor in the evaluation, we can account for these variances and obtain a more balanced view of a model's capabilities. This approach provides a representative assessment that mirrors real-world scenarios and emphasizes the importance of a model's adaptability and resilience in tackling challenges of varying complexity.

\section{Related Work}

Several benchmarks evaluating the coding capabilities of large language models exist. \citet{lu2021codexglue} categorize these evaluations into four main groups: code-to-code tasks (such as translation between programming languages or command line completion), text-to-text tasks (including documentation translation), code-to-text tasks (such as generating documentation for a function), and text-to-code tasks (involving instructions to generate a function for problem-solving).

The majority of well-known benchmarks primarily focus on code-to-code or text-to-code tasks. Benchmarks like HumanEval \citep{chen2021evaluating}, CoNaLa \citep{yin2018learning}, MMLU \citep{hendrycks2021measuring}, APPS \citep{2021_c24cd76e}, MBPP, MathQA \citep{austin2021program}, or HumanEvalPack \citep{muennighoff2023octopack} achieve this by presenting a diverse set of fundamental programming challenges. For instance, HumanEval evaluates the functional correctness of programs generated from docstrings and comprises 164 instances, while APPS encompasses both simple one-line problems and complex algorithmic challenges. Notably, these benchmarks lack realistic multi-turn settings where an agent engages with the generation system.

On the other hand, various mathematical benchmarks like GSM8K \cite{cobbe2021training}, and Rainbow \cite{Lourie2021UNICORNOR} do exist. However, its comprehensive coverage of difficulties, from basic to highly professional skill levels, sets Euler Project apart. This breadth enables a nuanced understanding of the specific tasks and their corresponding difficulty levels that current Large Language Models (LLMs) can handle effectively.

A multitude of benchmarks exist for measuring the reading comprehension of language models, including SQuAD \cite{rajpurkar2016squad}, CoQA \cite{reddy2019coqa}, LogiQA \cite{liu2020logiqa} or GLUE \cite{wang2019glue}. However, these benchmarks are different from the scope of our dataset. Our focus centers on solving complex tasks involving code generation, a skill achievable only with a specific reading comprehension level. Through our dataset, we contribute valuable insights into optimizing language model prompts. Specifically, we compare model performance between narrative-based and plain-math problems, enhancing our understanding of effectively engaging language models in challenging contexts.

\section{Conclusion}
We introduced the \datasetname{} dataset as a tool to assess the code generation capabilities of large language models across a spectrum of problem complexities, spanning both narrative and neutral contexts. Evaluations, which included a diverse set of models from proprietary to open-source and in-house, consistently highlighted the inherent challenges models face when navigating the intricacies of coding tasks, especially those embedded within narratives.

A significant observation from our study pertains to the evaluation formulation. The act of prompting in natural language, combined with the chat-based evaluation approach, introduces complexities. These complexities are both technical and conceptual, as they mirror real-world challenges where instructions and queries can be ambiguous or multifaceted. This underscores the need for models that can adeptly navigate the nuances of human language while maintaining computational accuracy.

Our comparative analysis between different problem subsets emphasized the differential impact of narrative versus neutral problem formulation on model performance spotlighted the potential for further exploration in this domain. Particularly, the Euler subset stood out, revealing substantial areas where models could improve code generation capabilities when faced with challenging math-based problem formulations.

Importantly, our methodology was rooted in a zero-shot prompting approach devoid of feedback loops. Transitioning to more sophisticated methodologies, such as chain-of-thought approaches \citep{wei2023chainofthought,le2023codechain} or tool-integrated reasoning-agents~\citep{gou2023tora}, hold promise for significantly enhancing model performance.

Our analysis uncovers a notable performance gap between commercial and open-source models. The introduced dataset proves valuable for evaluating models on coding and reasoning tasks, aiming to advance the assessment of LLMs in complex math and coding challenges. This work aims to enhance future research, expanding LLM capabilities in challenging fields.

\noindent{}
\textbf{Publishing Dataset}. \datasetname{} will be published over Github at \url{https://github.com/HallerPatrick/pecc} and 
HuggingFace at \url{https://huggingface.co/datasets/PatrickHaller/pecc}. The dataset will contain each problem, the respective input files, and expected solutions. The publication will not include the original AoC datasets due to licensing restrictions. Instead, we provide
a loading script to download the relevant datapoints.

\section{Acknowledgements}\
We thank all reviewers for their valuable comments. Alan Akbik and Patrick Haller are supported by the Deutsche Forschungsgemeinschaft (DFG, German Research Foundation) under Emmy Noether grant “Eidetic Representations of Natural Language” (project number 448414230).
Alan Akbik is furthermore supported under Germany’s Excellence Strategy "Science of Intelligence" (EXC 2002/1, project number 390523135).
We thank all reviewers for their valuable comments. 
Jonas Golde is supported by the German Federal Ministry of Economic Affairs and Climate Action (BMWK) as part of the project ENA (KK5148001LB0).

\section{Bibliographical References}\label{sec:reference}

\bibliographystyle{lrec-coling2024-natbib}
\bibliography{main}

\appendix

\end{document}

%% file: tables/comparison.tex
\begingroup
\renewcommand{\arraystretch}{1.3}
\begin{table*}[h]
    \centering
    \resizebox{0.99\textwidth}{!}{
    \begin{tabular}{lcccc}
    \toprule
    & APPS & HumanEval & DS-1000 &  \textbf{\datasetname{}} \\ 
    \midrule
    Programming\newline\ Language & Python & Python & Python & Universal \\
    Avg. Program\ Length & 18 & 6 & 3.6 & 26 (AoC) /\newline 19 (Euler)\\
    Number\ of\newline\ Problems & 10,000 & 164 & 1000 & 2352 \\
    Domain & Programming & Programming & Programming & Math \& Programming \\
    Evaluation & Test Cases & Python Code & \makecell{Test Cases +\\ Surface Form Constraints}  &  Explicit Result \\
    \bottomrule
    \end{tabular}}
    \caption{Comparison of Datasets with Textual Descriptions as Input. Unlike many existing datasets, \datasetname{} evaluates correctness through the explicit result assertion, making it universally applicable across various programming languages and problem-solving methodologies.}
    \label{tab:overview_subreddits}
\end{table*}
\endgroup

%% file: tables/result_table.tex
\begin{table*}[ht]
    \centering
\begin{threeparttable}
\def\arraystretch{1.2}
    \begin{tabular}{lcccccccc|c}
    \toprule
    Model & \multicolumn{2}{c}{AoC} & \multicolumn{2}{c}{AoC-Concise} & \multicolumn{2}{c}{Euler} & \multicolumn{2}{c}{Euler-Stories} & Average\\ 
    & $k = 1$ & $k = 3$ & $k = 1$ & $k = 3$ & $k = 1$ & $k = 3$ & $k = 1$ & $k = 3$ & $k = 3$ \\
    \midrule
    claude-haiku\tnote{a} & 37.76 & 51.28 & 41.33 & 46.26 & 5.58 & 7.07 & 4.59 & 6.08 & 27.67 \\
    gpt-3.5-turbo\tnote{a} & 38.52 & 50.00 & 28.32 & 29.85 & 7.20& 8.19 & 5.33 & 6.95 & 23.75 \\
    codechat-bison\tnote{a} & 12.76& 21.17 & 15.05 & 17.60 & 3.47 & 4.59 & 2.11 & 2.61 & 11.39 \\
    chat-bison\tnote{a} & 16.07 & 17.09 & 14.03 & 13.78 & 2.36 & 2.36 & 0.62 & 0.62 & 8.48 \\
    Mixtral-8x7B-Instruct & 8.93 & 15.31 & 5.36 & 13.01 & 1.86 & 2.86 & 0.75 & 2.23 & 8.35 \\
    Phi-3-mini-128k-instruct & 9.13 & 10.13 & 9.95 & 13.00 & 0.0 & 3.35 & 1.24 & 2.23 & 7.18 \\
    WizardLM-2-7b & 4.57 & 5.87 & 6.33 & 6.89 & 0.37 & 1.24 & 0.37 & 0.87 & 3.72 \\
    Llama-3-8B-Instruct & 1.46 & 1.53 & 5.50 & 6.38 & 2.73 & 4.47 & 0.0 & 0.0 & 3.1 \\
    \hdashline
    WizardCoder-Python\tnote{b} & 11.00 & 24.00 & 9.50 & 22.50 & 1.49 & 2.61 & 1.36 & 2.48 & 12.9 \\
    Mistral-7B-Instruct\tnote{b} & 3.00 & 3.00 & 3.00 & 3.00 & 0.3 & 0.3 & 0.12 & 0.12 & 1.62  \\ 
    \midrule
    \bottomrule
    \end{tabular}
    \begin{tablenotes} \small
        \item[a] We note that we could confirm that the models have seen the original AoC Challenges during pre-training, as it is able to generate a challenge for a given year and day.
        \item[b] Excludes evaluation on second part of AoC subsets.
    \end{tablenotes}
    \caption{Pass@k evaluation of different proprietary and open-source models.}
    \label{tab:output_evaluation}
\end{threeparttable}
\end{table*}

%% file: tables/error_types.tex
\begin{table*}
\centering
\begin{tabular}{llccccr}
\toprule
& & \multicolumn{4}{c}{Error Types (in \%)} & \\
Model & Subset & Syntax & Runtime & Timeout & Wrong & Total Errors \\
\midrule
\multirow{4}{*}{gpt-3.5-turbo} & AoC & \cellcolor{color4} 20.2 & \cellcolor{color9} 45.4 & \cellcolor{color1} 6.7 & \cellcolor{color5} 27.7 & 119 \\
& AoC-Leet & \cellcolor{color1} 5.9 & \cellcolor{color7} 36.6 & \cellcolor{color1} 9.2 & \cellcolor{color9} 48.4 & 153 \\
& Euler & \cellcolor{color0} 0.8 & \cellcolor{color2} 12.3 & \cellcolor{color6} 30.1 & \cellcolor{color11} 56.8 & 740 \\
& Euler-Stories & \cellcolor{color0} 1.9 & \cellcolor{color2} 11.2 & \cellcolor{color5} 26.5 & \cellcolor{color12} 60.4 & 750 \\
\midrule
\multirow{4}{*}{claude-haiku} & AoC & \cellcolor{color1} 7.2 & \cellcolor{color7} 38.7 & \cellcolor{color2} 11.7 & \cellcolor{color8} 42.3 & 111 \\
& AoC-Leet & \cellcolor{color0} 3.8 & \cellcolor{color8} 44.9 & \cellcolor{color2} 14.6 & \cellcolor{color7} 36.7 & 158 \\
& Euler & \cellcolor{color0} 1.3 & \cellcolor{color3} 16.0 & \cellcolor{color10} 51.0 & \cellcolor{color6} 31.6 & 749 \\
& Euler-Stories & \cellcolor{color0} 1.1 & \cellcolor{color2} 12.5 & \cellcolor{color7} 37.3 & \cellcolor{color9} 49.1 & 757 \\
\midrule
\multirow{4}{*}{chat-bison} & AoC & \cellcolor{color0} 3.4 & \cellcolor{color14} 72.7 & \cellcolor{color1} 6.2 & \cellcolor{color3} 17.6 & 176 \\
& AoC-Leet & \cellcolor{color0} 0.5 & \cellcolor{color14} 73.4 & \cellcolor{color1} 6.0 & \cellcolor{color4} 20.1 & 184 \\
& Euler & \cellcolor{color0} 0.4 & \cellcolor{color8} 40.4 & \cellcolor{color4} 20.6 & \cellcolor{color6} 34.9 & 787 \\
& Euler-Stories & \cellcolor{color0} 0.6 & \cellcolor{color8} 42.3 & \cellcolor{color2} 14.1 & \cellcolor{color8} 42.9 & 801 \\
\midrule
\multirow{4}{*}{codechat-bison} & AoC & \cellcolor{color0} 0.6 & \cellcolor{color15} 75.4 & \cellcolor{color1} 5.7 & \cellcolor{color3} 18.3 &  175\\
& AoC-Leet & \cellcolor{color0} 0.6 & \cellcolor{color14} 74.2 & \cellcolor{color1} 6.2 & \cellcolor{color3} 19.1 & 178 \\
& Euler & \cellcolor{color0} 0.1 & \cellcolor{color7} 39.8 & \cellcolor{color5} 26.4 & \cellcolor{color6} 33.7 & 769 \\
& Euler-Stories & \cellcolor{color0} 0.1 & \cellcolor{color7} 39.5 & \cellcolor{color4} 20.1 & \cellcolor{color8} 40.3 & 785 \\
\midrule
\multirow{4}{*}{wizard-coder} & AoC & \cellcolor{color5} 27.6 & \cellcolor{color6} 30.3 & \cellcolor{color1} 9.9 & \cellcolor{color6} 32.2 & 152 \\
& AoC-Leet & \cellcolor{color5} 26.5 & \cellcolor{color6} 32.3 & \cellcolor{color2} 12.9 & \cellcolor{color5} 28.4 & 155 \\
& Euler & \cellcolor{color2} 13.0 & \cellcolor{color5} 27.0 & \cellcolor{color5} 29.9 & \cellcolor{color6} 30.1 & 785 \\
& Euler-Stories & \cellcolor{color3} 15.1 & \cellcolor{color5} 25.4 & \cellcolor{color5} 28.4 & \cellcolor{color6} 31.0 & 786 \\
\midrule
\multirow{4}{*}{mistral-instruct} & AoC & \cellcolor{color1} 5.2 & \cellcolor{color12} 62.4 & \cellcolor{color1} 8.8 & \cellcolor{color4} 23.7 & 194 \\
& AoC-Leet & \cellcolor{color1} 5.2 & \cellcolor{color11} 58.2 & \cellcolor{color2} 10.8 & \cellcolor{color5} 25.8 & 194 \\
& Euler & \cellcolor{color1} 6.1 & \cellcolor{color8} 43.8 & \cellcolor{color2} 13.0 & \cellcolor{color7} 37.1 & 803 \\
& Euler-Stories & \cellcolor{color1} 5.3 & \cellcolor{color9} 45.6 & \cellcolor{color1} 8.1 & \cellcolor{color8} 41.0 & 805 \\

\bottomrule
\end{tabular}
\caption{Overview over different error types that led to failing a problem in Pass@3 evaluation.}
\label{tab:error_types}
\end{table*}

%% file: main.bbl
\begin{thebibliography}{26}
\expandafter\ifx\csname natexlab\endcsname\relax\def\natexlab#1{#1}\fi

\bibitem[{Abdin et~al.(2024)Abdin, Jacobs, Awan, Aneja, Awadallah, Awadalla, Bach, Bahree, Bakhtiari, Behl, Benhaim, Bilenko, Bjorck, Bubeck, Cai, Mendes, Chen, Chaudhary, Chopra, Giorno, de~Rosa, Dixon, Eldan, Iter, Garg, Goswami, Gunasekar, Haider, Hao, Hewett, Huynh, Javaheripi, Jin, Kauffmann, Karampatziakis, Kim, Khademi, Kurilenko, Lee, Lee, Li, Liang, Liu, Lin, Lin, Madan, Mitra, Modi, Nguyen, Norick, Patra, Perez-Becker, Portet, Pryzant, Qin, Radmilac, Rosset, Roy, Ruwase, Saarikivi, Saied, Salim, Santacroce, Shah, Shang, Sharma, Song, Tanaka, Wang, Ward, Wang, Witte, Wyatt, Xu, Xu, Yadav, Yang, Yang, Yu, Zhang, Zhang, Zhang, Zhang, Zhang, Zhang, Zhang, and Zhou}]{abdin2024phi3}
Marah Abdin, Sam~Ade Jacobs, Ammar~Ahmad Awan, Jyoti Aneja, Ahmed Awadallah, Hany Awadalla, Nguyen Bach, Amit Bahree, Arash Bakhtiari, Harkirat Behl, Alon Benhaim, Misha Bilenko, Johan Bjorck, Sébastien Bubeck, Martin Cai, Caio César~Teodoro Mendes, Weizhu Chen, Vishrav Chaudhary, Parul Chopra, Allie~Del Giorno, Gustavo de~Rosa, Matthew Dixon, Ronen Eldan, Dan Iter, Amit Garg, Abhishek Goswami, Suriya Gunasekar, Emman Haider, Junheng Hao, Russell~J. Hewett, Jamie Huynh, Mojan Javaheripi, Xin Jin, Piero Kauffmann, Nikos Karampatziakis, Dongwoo Kim, Mahoud Khademi, Lev Kurilenko, James~R. Lee, Yin~Tat Lee, Yuanzhi Li, Chen Liang, Weishung Liu, Eric Lin, Zeqi Lin, Piyush Madan, Arindam Mitra, Hardik Modi, Anh Nguyen, Brandon Norick, Barun Patra, Daniel Perez-Becker, Thomas Portet, Reid Pryzant, Heyang Qin, Marko Radmilac, Corby Rosset, Sambudha Roy, Olatunji Ruwase, Olli Saarikivi, Amin Saied, Adil Salim, Michael Santacroce, Shital Shah, Ning Shang, Hiteshi Sharma, Xia Song, Masahiro Tanaka, Xin Wang, Rachel
  Ward, Guanhua Wang, Philipp Witte, Michael Wyatt, Can Xu, Jiahang Xu, Sonali Yadav, Fan Yang, Ziyi Yang, Donghan Yu, Chengruidong Zhang, Cyril Zhang, Jianwen Zhang, Li~Lyna Zhang, Yi~Zhang, Yue Zhang, Yunan Zhang, and Xiren Zhou. 2024.
\newblock \href {http://arxiv.org/abs/2404.14219} {Phi-3 technical report: A highly capable language model locally on your phone}.

\bibitem[{AI@Meta(2024)}]{llama3modelcard}
AI@Meta. 2024.
\newblock \href {https://github.com/meta-llama/llama3/blob/main/MODEL_CARD.md} {Llama 3 model card}.

\bibitem[{Anil et~al.(2023)Anil, Dai, Firat, Johnson, Lepikhin, Passos, Shakeri, Taropa, Bailey, Chen, Chu, Clark, Shafey, Huang, Meier-Hellstern, Mishra, Moreira, Omernick, Robinson, Ruder, Tay, Xiao, Xu, Zhang, Abrego, Ahn, Austin, Barham, Botha, Bradbury, Brahma, Brooks, Catasta, Cheng, Cherry, Choquette-Choo, Chowdhery, Crepy, Dave, Dehghani, Dev, Devlin, Díaz, Du, Dyer, Feinberg, Feng, Fienber, Freitag, Garcia, Gehrmann, Gonzalez, Gur-Ari, Hand, Hashemi, Hou, Howland, Hu, Hui, Hurwitz, Isard, Ittycheriah, Jagielski, Jia, Kenealy, Krikun, Kudugunta, Lan, Lee, Lee, Li, Li, Li, Li, Li, Lim, Lin, Liu, Liu, Maggioni, Mahendru, Maynez, Misra, Moussalem, Nado, Nham, Ni, Nystrom, Parrish, Pellat, Polacek, Polozov, Pope, Qiao, Reif, Richter, Riley, Ros, Roy, Saeta, Samuel, Shelby, Slone, Smilkov, So, Sohn, Tokumine, Valter, Vasudevan, Vodrahalli, Wang, Wang, Wang, Wang, Wieting, Wu, Xu, Xu, Xue, Yin, Yu, Zhang, Zheng, Zheng, Zhou, Zhou, Petrov, and Wu}]{anil2023palm}
Rohan Anil, Andrew~M. Dai, Orhan Firat, Melvin Johnson, Dmitry Lepikhin, Alexandre Passos, Siamak Shakeri, Emanuel Taropa, Paige Bailey, Zhifeng Chen, Eric Chu, Jonathan~H. Clark, Laurent~El Shafey, Yanping Huang, Kathy Meier-Hellstern, Gaurav Mishra, Erica Moreira, Mark Omernick, Kevin Robinson, Sebastian Ruder, Yi~Tay, Kefan Xiao, Yuanzhong Xu, Yujing Zhang, Gustavo~Hernandez Abrego, Junwhan Ahn, Jacob Austin, Paul Barham, Jan Botha, James Bradbury, Siddhartha Brahma, Kevin Brooks, Michele Catasta, Yong Cheng, Colin Cherry, Christopher~A. Choquette-Choo, Aakanksha Chowdhery, Clément Crepy, Shachi Dave, Mostafa Dehghani, Sunipa Dev, Jacob Devlin, Mark Díaz, Nan Du, Ethan Dyer, Vlad Feinberg, Fangxiaoyu Feng, Vlad Fienber, Markus Freitag, Xavier Garcia, Sebastian Gehrmann, Lucas Gonzalez, Guy Gur-Ari, Steven Hand, Hadi Hashemi, Le~Hou, Joshua Howland, Andrea Hu, Jeffrey Hui, Jeremy Hurwitz, Michael Isard, Abe Ittycheriah, Matthew Jagielski, Wenhao Jia, Kathleen Kenealy, Maxim Krikun, Sneha Kudugunta, Chang
  Lan, Katherine Lee, Benjamin Lee, Eric Li, Music Li, Wei Li, YaGuang Li, Jian Li, Hyeontaek Lim, Hanzhao Lin, Zhongtao Liu, Frederick Liu, Marcello Maggioni, Aroma Mahendru, Joshua Maynez, Vedant Misra, Maysam Moussalem, Zachary Nado, John Nham, Eric Ni, Andrew Nystrom, Alicia Parrish, Marie Pellat, Martin Polacek, Alex Polozov, Reiner Pope, Siyuan Qiao, Emily Reif, Bryan Richter, Parker Riley, Alex~Castro Ros, Aurko Roy, Brennan Saeta, Rajkumar Samuel, Renee Shelby, Ambrose Slone, Daniel Smilkov, David~R. So, Daniel Sohn, Simon Tokumine, Dasha Valter, Vijay Vasudevan, Kiran Vodrahalli, Xuezhi Wang, Pidong Wang, Zirui Wang, Tao Wang, John Wieting, Yuhuai Wu, Kelvin Xu, Yunhan Xu, Linting Xue, Pengcheng Yin, Jiahui Yu, Qiao Zhang, Steven Zheng, Ce~Zheng, Weikang Zhou, Denny Zhou, Slav Petrov, and Yonghui Wu. 2023.
\newblock \href {http://arxiv.org/abs/2305.10403} {Palm 2 technical report}.

\bibitem[{Anthropic(2024)}]{claude-3}
Anthropic. 2024.
\newblock \href {https://www.anthropic.com/news/claude-3-family} {The claude 3 model family: Opus, sonnet, haiku}.

\bibitem[{Austin et~al.(2021)Austin, Odena, Nye, Bosma, Michalewski, Dohan, Jiang, Cai, Terry, Le, and Sutton}]{austin2021program}
Jacob Austin, Augustus Odena, Maxwell Nye, Maarten Bosma, Henryk Michalewski, David Dohan, Ellen Jiang, Carrie Cai, Michael Terry, Quoc Le, and Charles Sutton. 2021.
\newblock \href {http://arxiv.org/abs/2108.07732} {Program synthesis with large language models}.

\bibitem[{Chase(2022)}]{langchain}
Harrison Chase. 2022.
\newblock \href {https://github.com/langchain-ai/langchain} {Langchain}.

\bibitem[{Chen et~al.(2021)Chen, Tworek, Jun, Yuan, de~Oliveira~Pinto, Kaplan, Edwards, Burda, Joseph, Brockman, Ray, Puri, Krueger, Petrov, Khlaaf, Sastry, Mishkin, Chan, Gray, Ryder, Pavlov, Power, Kaiser, Bavarian, Winter, Tillet, Such, Cummings, Plappert, Chantzis, Barnes, Herbert-Voss, Guss, Nichol, Paino, Tezak, Tang, Babuschkin, Balaji, Jain, Saunders, Hesse, Carr, Leike, Achiam, Misra, Morikawa, Radford, Knight, Brundage, Murati, Mayer, Welinder, McGrew, Amodei, McCandlish, Sutskever, and Zaremba}]{chen2021evaluating}
Mark Chen, Jerry Tworek, Heewoo Jun, Qiming Yuan, Henrique~Ponde de~Oliveira~Pinto, Jared Kaplan, Harri Edwards, Yuri Burda, Nicholas Joseph, Greg Brockman, Alex Ray, Raul Puri, Gretchen Krueger, Michael Petrov, Heidy Khlaaf, Girish Sastry, Pamela Mishkin, Brooke Chan, Scott Gray, Nick Ryder, Mikhail Pavlov, Alethea Power, Lukasz Kaiser, Mohammad Bavarian, Clemens Winter, Philippe Tillet, Felipe~Petroski Such, Dave Cummings, Matthias Plappert, Fotios Chantzis, Elizabeth Barnes, Ariel Herbert-Voss, William~Hebgen Guss, Alex Nichol, Alex Paino, Nikolas Tezak, Jie Tang, Igor Babuschkin, Suchir Balaji, Shantanu Jain, William Saunders, Christopher Hesse, Andrew~N. Carr, Jan Leike, Josh Achiam, Vedant Misra, Evan Morikawa, Alec Radford, Matthew Knight, Miles Brundage, Mira Murati, Katie Mayer, Peter Welinder, Bob McGrew, Dario Amodei, Sam McCandlish, Ilya Sutskever, and Wojciech Zaremba. 2021.
\newblock \href {http://arxiv.org/abs/2107.03374} {Evaluating large language models trained on code}.

\bibitem[{Cobbe et~al.(2021)Cobbe, Kosaraju, Bavarian, Chen, Jun, Kaiser, Plappert, Tworek, Hilton, Nakano, Hesse, and Schulman}]{cobbe2021training}
Karl Cobbe, Vineet Kosaraju, Mohammad Bavarian, Mark Chen, Heewoo Jun, Lukasz Kaiser, Matthias Plappert, Jerry Tworek, Jacob Hilton, Reiichiro Nakano, Christopher Hesse, and John Schulman. 2021.
\newblock \href {http://arxiv.org/abs/2110.14168} {Training verifiers to solve math word problems}.

\bibitem[{Gou et~al.(2023)Gou, Shao, Gong, yelong shen, Yang, Huang, Duan, and Chen}]{gou2023tora}
Zhibin Gou, Zhihong Shao, Yeyun Gong, yelong shen, Yujiu Yang, Minlie Huang, Nan Duan, and Weizhu Chen. 2023.
\newblock \href {http://arxiv.org/abs/2309.17452} {Tora: A tool-integrated reasoning agent for mathematical problem solving}.

\bibitem[{Hendrycks et~al.(2021{\natexlab{a}})Hendrycks, Basart, Kadavath, Mazeika, Arora, Guo, Burns, Puranik, He, Song, and Steinhardt}]{2021_c24cd76e}
Dan Hendrycks, Steven Basart, Saurav Kadavath, Mantas Mazeika, Akul Arora, Ethan Guo, Collin Burns, Samir Puranik, Horace He, Dawn Song, and Jacob Steinhardt. 2021{\natexlab{a}}.
\newblock \href {https://datasets-benchmarks-proceedings.neurips.cc/paper_files/paper/2021/file/c24cd76e1ce41366a4bbe8a49b02a028-Paper-round2.pdf} {Measuring coding challenge competence with apps}.
\newblock In \emph{Proceedings of the Neural Information Processing Systems Track on Datasets and Benchmarks}, volume~1. Curran.

\bibitem[{Hendrycks et~al.(2021{\natexlab{b}})Hendrycks, Burns, Basart, Zou, Mazeika, Song, and Steinhardt}]{hendrycks2021measuring}
Dan Hendrycks, Collin Burns, Steven Basart, Andy Zou, Mantas Mazeika, Dawn Song, and Jacob Steinhardt. 2021{\natexlab{b}}.
\newblock \href {http://arxiv.org/abs/2009.03300} {Measuring massive multitask language understanding}.

\bibitem[{Jiang et~al.(2023)Jiang, Sablayrolles, Mensch, Bamford, Chaplot, de~las Casas, Bressand, Lengyel, Lample, Saulnier, Lavaud, Lachaux, Stock, Scao, Lavril, Wang, Lacroix, and Sayed}]{jiang2023mistral}
Albert~Q. Jiang, Alexandre Sablayrolles, Arthur Mensch, Chris Bamford, Devendra~Singh Chaplot, Diego de~las Casas, Florian Bressand, Gianna Lengyel, Guillaume Lample, Lucile Saulnier, Lélio~Renard Lavaud, Marie-Anne Lachaux, Pierre Stock, Teven~Le Scao, Thibaut Lavril, Thomas Wang, Timothée Lacroix, and William~El Sayed. 2023.
\newblock \href {http://arxiv.org/abs/2310.06825} {Mistral 7b}.

\bibitem[{Jiang et~al.(2024)Jiang, Sablayrolles, Roux, Mensch, Savary, Bamford, Chaplot, de~las Casas, Hanna, Bressand, Lengyel, Bour, Lample, Lavaud, Saulnier, Lachaux, Stock, Subramanian, Yang, Antoniak, Scao, Gervet, Lavril, Wang, Lacroix, and Sayed}]{jiang2024mixtral}
Albert~Q. Jiang, Alexandre Sablayrolles, Antoine Roux, Arthur Mensch, Blanche Savary, Chris Bamford, Devendra~Singh Chaplot, Diego de~las Casas, Emma~Bou Hanna, Florian Bressand, Gianna Lengyel, Guillaume Bour, Guillaume Lample, Lélio~Renard Lavaud, Lucile Saulnier, Marie-Anne Lachaux, Pierre Stock, Sandeep Subramanian, Sophia Yang, Szymon Antoniak, Teven~Le Scao, Théophile Gervet, Thibaut Lavril, Thomas Wang, Timothée Lacroix, and William~El Sayed. 2024.
\newblock \href {http://arxiv.org/abs/2401.04088} {Mixtral of experts}.

\bibitem[{Lai et~al.(2022)Lai, Li, Wang, Zhang, Zhong, Zettlemoyer, tau Yih, Fried, Wang, and Yu}]{lai2022ds1000}
Yuhang Lai, Chengxi Li, Yiming Wang, Tianyi Zhang, Ruiqi Zhong, Luke Zettlemoyer, Scott~Wen tau Yih, Daniel Fried, Sida Wang, and Tao Yu. 2022.
\newblock \href {http://arxiv.org/abs/2211.11501} {Ds-1000: A natural and reliable benchmark for data science code generation}.

\bibitem[{Le et~al.(2023)Le, Chen, Saha, Gokul, Sahoo, and Joty}]{le2023codechain}
Hung Le, Hailin Chen, Amrita Saha, Akash Gokul, Doyen Sahoo, and Shafiq Joty. 2023.
\newblock \href {http://arxiv.org/abs/2310.08992} {Codechain: Towards modular code generation through chain of self-revisions with representative sub-modules}.

\bibitem[{Liu et~al.(2020)Liu, Cui, Liu, Huang, Wang, and Zhang}]{liu2020logiqa}
Jian Liu, Leyang Cui, Hanmeng Liu, Dandan Huang, Yile Wang, and Yue Zhang. 2020.
\newblock \href {http://arxiv.org/abs/2007.08124} {Logiqa: A challenge dataset for machine reading comprehension with logical reasoning}.

\bibitem[{Lourie et~al.(2021)Lourie, Bras, Bhagavatula, and Choi}]{Lourie2021UNICORNOR}
Nicholas Lourie, Ronan~Le Bras, Chandra Bhagavatula, and Yejin Choi. 2021.
\newblock \href {https://api.semanticscholar.org/CorpusID:232335877} {Unicorn on rainbow: A universal commonsense reasoning model on a new multitask benchmark}.
\newblock \emph{ArXiv}, abs/2103.13009.

\bibitem[{Lu et~al.(2021)Lu, Guo, Ren, Huang, Svyatkovskiy, Blanco, Clement, Drain, Jiang, Tang, Li, Zhou, Shou, Zhou, Tufano, Gong, Zhou, Duan, Sundaresan, Deng, Fu, and Liu}]{lu2021codexglue}
Shuai Lu, Daya Guo, Shuo Ren, Junjie Huang, Alexey Svyatkovskiy, Ambrosio Blanco, Colin Clement, Dawn Drain, Daxin Jiang, Duyu Tang, Ge~Li, Lidong Zhou, Linjun Shou, Long Zhou, Michele Tufano, Ming Gong, Ming Zhou, Nan Duan, Neel Sundaresan, Shao~Kun Deng, Shengyu Fu, and Shujie Liu. 2021.
\newblock \href {http://arxiv.org/abs/2102.04664} {Codexglue: A machine learning benchmark dataset for code understanding and generation}.

\bibitem[{Luo et~al.(2023)Luo, Xu, Zhao, Sun, Geng, Hu, Tao, Ma, Lin, and Jiang}]{luo2023wizardcoder}
Ziyang Luo, Can Xu, Pu~Zhao, Qingfeng Sun, Xiubo Geng, Wenxiang Hu, Chongyang Tao, Jing Ma, Qingwei Lin, and Daxin Jiang. 2023.
\newblock Wizardcoder: Empowering code large language models with evol-instruct.
\newblock \emph{arXiv preprint arXiv:2306.08568}.

\bibitem[{Muennighoff et~al.(2023)Muennighoff, Liu, Zebaze, Zheng, Hui, Zhuo, Singh, Tang, von Werra, and Longpre}]{muennighoff2023octopack}
Niklas Muennighoff, Qian Liu, Armel Zebaze, Qinkai Zheng, Binyuan Hui, Terry~Yue Zhuo, Swayam Singh, Xiangru Tang, Leandro von Werra, and Shayne Longpre. 2023.
\newblock \href {http://arxiv.org/abs/2308.07124} {Octopack: Instruction tuning code large language models}.

\bibitem[{OpenAI(2023)}]{OpenAI2023GPT4TR}
OpenAI. 2023.
\newblock \href {https://api.semanticscholar.org/CorpusID:257532815} {Gpt-4 technical report}.
\newblock \emph{ArXiv}, abs/2303.08774.

\bibitem[{Rajpurkar et~al.(2016)Rajpurkar, Zhang, Lopyrev, and Liang}]{rajpurkar2016squad}
Pranav Rajpurkar, Jian Zhang, Konstantin Lopyrev, and Percy Liang. 2016.
\newblock \href {http://arxiv.org/abs/1606.05250} {Squad: 100,000+ questions for machine comprehension of text}.

\bibitem[{Reddy et~al.(2019)Reddy, Chen, and Manning}]{reddy2019coqa}
Siva Reddy, Danqi Chen, and Christopher~D. Manning. 2019.
\newblock \href {http://arxiv.org/abs/1808.07042} {Coqa: A conversational question answering challenge}.

\bibitem[{Wang et~al.(2019)Wang, Singh, Michael, Hill, Levy, and Bowman}]{wang2019glue}
Alex Wang, Amanpreet Singh, Julian Michael, Felix Hill, Omer Levy, and Samuel~R. Bowman. 2019.
\newblock \href {http://arxiv.org/abs/1804.07461} {Glue: A multi-task benchmark and analysis platform for natural language understanding}.

\bibitem[{Wei et~al.(2023)Wei, Wang, Schuurmans, Bosma, Ichter, Xia, Chi, Le, and Zhou}]{wei2023chainofthought}
Jason Wei, Xuezhi Wang, Dale Schuurmans, Maarten Bosma, Brian Ichter, Fei Xia, Ed~Chi, Quoc Le, and Denny Zhou. 2023.
\newblock \href {http://arxiv.org/abs/2201.11903} {Chain-of-thought prompting elicits reasoning in large language models}.

\bibitem[{Yin et~al.(2018)Yin, Deng, Chen, Vasilescu, and Neubig}]{yin2018learning}
Pengcheng Yin, Bowen Deng, Edgar Chen, Bogdan Vasilescu, and Graham Neubig. 2018.
\newblock \href {http://arxiv.org/abs/1805.08949} {Learning to mine aligned code and natural language pairs from stack overflow}.

\end{thebibliography}
